\documentclass[runningheads]{llncs}
\usepackage{caption}
\usepackage{subcaption}
\usepackage{xcolor}
\usepackage[T1]{fontenc}
\usepackage{multirow}
\usepackage{graphicx}
\usepackage{braket}
\begin{document}
\title
{An ensemble framework approach of hybrid Quantum convolutional neural networks for classification of breast cancer images}
%
%
\author{Dibyasree Guha\inst{1}\orcidID{0009-0003-4651-9297} \and Shyamali Mitra\inst{2}\orcidID{0000-0001-6502-9056} \and
Somenath Kuiry\inst{3,4}\orcidID{0000-0002-8462-5547} \and
Nibaran Das\inst{1}\orcidID{0000-0002-2426-9915 }}
\authorrunning{{Guha D.} et al.}
%
\institute{Department of CSE, Jadavpur University, Kolkata \\
\and Department of IEE, Jadavpur University, Kolkata \\
\and Department of Mathematics, Jadavpur University, Kolkata \\
\and Department of AIML, Institute of Engineering \& Management, Kolkata
\email{\{dibyasrees.cse.rs,shyamalimitra.iee,skuiry.math.rs,nibaran.das\}@jadavpuruniversity.in}}
\maketitle              
\begin{abstract}
Quantum neural networks are deemed suitable to replace classical neural networks in their ability to learn and scale up network models using quantum-exclusive phenomena like superposition and entanglement. However, in the noisy intermediate scale quantum (NISQ) era, the trainability and expressibility of quantum models are yet under investigation. Medical image classification on the other hand, pertains well to applications in deep learning, particularly, convolutional neural networks.  In this paper, we carry out a study of three hybrid classical-quantum neural network architectures and combine them using standard ensembling techniques on a breast cancer histopathological dataset. The best accuracy percentage obtained by an individual model is $85.59$. Whereas, on performing ensemble, we have obtained accuracy as high as $86.72\%$, an improvement over the individual hybrid network as well as classical neural network counterparts of the hybrid network models.

\end{abstract}
%
%
%
\section{Introduction}
Research in the field of machine learning is experiencing advances in leaps and bounds in recent times, especially in the domain of medical image analysis. However, with the recent turn of events in the domain of quantum computation, quantum tools to boost machine learning operations are beginning to gain a lot of attention \cite{rebentrost2014quantum,Xin2021ExperimentalQP,Zhou}. Variational quantum classifiers - quantum circuits that can be trained from labeled data to classify new data samples, stand out in this regard. The quantum circuits are essentially characterized by varying certain parameter values, hence the name parameterized quantum circuit (PQC).



An emerging field in the research domain unifies quantum mechanics concepts with enhanced computational capabilities. The consequence of this amalgamation has produced the principle of quantum-classical hybridization that requires investigation in the wake of the NISQ era. A quantum neural network(QNN) is defined in various ways in the literature, but it can generally be understood as a type of variational quantum circuit that consists of adjustable parameters. Essentially, a QNN is a variational quantum circuit whose parameters i.e. qubit rotations, can be optimized through training. The measurement outcomes of this circuit aim to approximate the target, such as the label for a machine learning task \cite{abbas2021power}. Variational quantum algorithms contain parameterized quantum circuits that consist of parameter-driven quantum gates \cite{schuld2020circuit}. The quantum gates need to be fed information in the form of quantum states. Thus, state preparation is the foremost task to encode classical information. The encoded information is trained using the parameterized circuit, that derives the value of its parameters from the outputs of a classical subroutine. The training involves optimization for the task at hand through gradient descent, which in turn requires gradient computation of the parameterized quantum circuit \cite{crooks2019gradients}. Measurement of the output of the quantum circuit is processed with the help of classical techniques to yield the network outcome. The parameterized circuits begin with an ansatz, which can be regarded as an educated first guess, which we then optimize through repeated iterations of the circuit. Quantum variational classifiers operate in higher effective dimensions and possess faster training capabilities, making them superior candidates compared to classical models.
However, experiments with pure quantum neural networks are not trivial with traditional computers, and procurement of quantum hardware is difficult. Quantum simulators have the deficiency in terms of qubits and cannot handle images of larger dimensions. In this paper, we explore the possibility of using hybrid models which utilize an amalgamation of both classical and quantum frameworks, for the task of medical image classification. Hybrid networks are simple to comprehend and implementation is fairly easy. In such models, the feature extraction is done by a classical model, and the final class probabilities are calculated by a parameterized quantum circuit using rotation gates. In this work, we have created three different hybrid QCNN models each having a different network structure than the other, the classification performances of which are fairly satisfactory. To further enhance the performance mentioned above, we have employed a few ensemble techniques as well as combining the individual hybrid network outputs.
The subsequent sections are organized as follows: previous work on PQCs and hybrid quantum-classical deep neural networks have been discussed in section 2. The three different hybrid network architectures have been elaborated in section 3, followed by section 4 which contains the dataset description along with details of the experimental procedures involving ensemble approaches. Section 5 involves the performance figures of individual hybrid models as well as those acquired on the application of ensembling methods. Section 6 is dedicated to the concluding remarks and future scope.

\section{Previous Work}
 PQCs are often realized in quantum-classical hybridization mode. In 2014, a classical computer and a photon quantum processor jointly contributed to building a variational quantum eigensolver (VQE) - a method proposed by Peruzzo et al \cite{Peruzzo}. Another work that contributed to improving the previously-mentioned VQE scheme was that by McClean et al. \cite{Mcclean} in 2016. A quantum classifier showing promising performance measures was built by Schuld et al. \cite{schuld2020circuit}. A PQC-based hybrid quantum neural network (QNN) was proposed by Zeng et al. \cite{Zeng}. It involved a ladder circuit structure. Combining classical convolutional neural networks with parameterized quantum circuits paved the way for quantum convolutional neural networks (QCNN). The first of its kind was presented by Cong et al. \cite{Cong} in 2019. The QCNN model, suitable enough to be trained and implemented in NISQ devices efficiently, was used for the applications of phase classification and quantum error correction codes optimization. In 2020, Henderson et al. \cite{Henderson} demonstrated that the introduction of quantum convolution or quanvolution layers resulted in higher test accuracies coupled with faster training than purely classical CNNs. In 2020, Li et al. \cite{Li} introduced a quantum analogue of classical convolution—a Quantum Deep Convolutional Neural Network (QDCNN)—which demonstrated exponential computational acceleration. Following this, in 2021, Parthasarathy et al. \cite{9492087} developed a Quantum Optical Convolutional Neural Network (QOCNN) that seamlessly integrated quantum convolution with quantum optics. The network
exhibited good stability in terms of image recognition.
In 2021, Liu et al. \cite{Liu} proposed an NISQ-friendly hybrid quantum-classical convolutional neural network (QCCNN) and applied it to a Tetris dataset that performed classification tasks with better learning accuracy than a classical CNN having a similar structure. In 2022, Wei et al. \cite{Wei} came up with a robust QCNN for image recognition and image processing tasks that also demonstrated reduced computational complexity than purely classical CNN. In 2023, Fan et al. \cite{Fan} introduced a hybrid quantum-classical CNN (QC-CNN) that leverages quantum computing to effectively extract important high-level features from Earth Observation data for image classification purposes.

\section{Hybrid Neural Network Architecture}
\begin{figure}
\includegraphics[width=\textwidth]{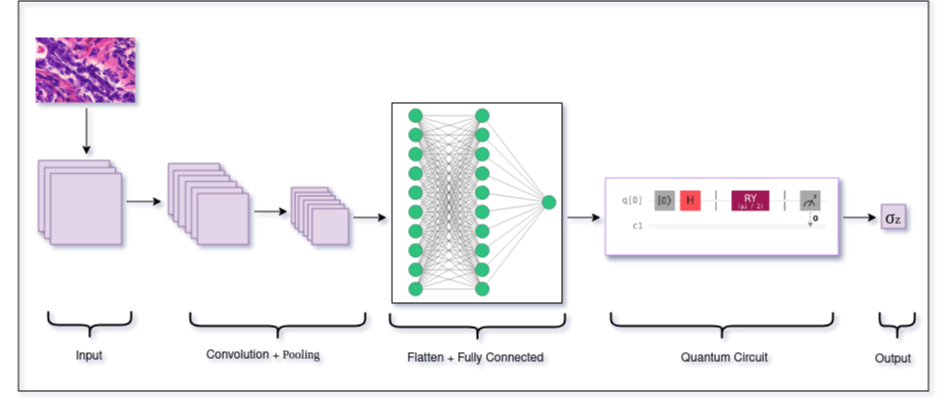}
\caption{Hybrid Classical-Quantum Neural Network} \label{fig0}
\end{figure}

 This section describes the construction of hybrid quantum-classical neural network architectures that can be used for classification tasks using quantum machine learning. The hybrid neural network architecture consists of two parts - a classical feature extractor part and a quantum gate to produce final class probabilities. The entire framework can be considered to have an input layer, a hidden layer, and an output layer. The first and the last layers are classical in nature while the hidden layer implements the quantum component of our model in the form of a Parametrized Quantum Circuit (PQC). The PQC is a quantum circuit where rotation angles for each gate are specified using components from a classical input vector. The underlying model comprises a classical convolutional neural network (CNN) used as a feature extractor to produce a low-dimensional feature vector, which is subsequently fed into a parameterized quantum circuit for classification. For the quantum part of these hybrid(quantum-classical) networks, we created a quantum circuit and then vary the parameters to minimize the objective function of interest using the parameter shift rule \cite{crooks2019gradients}. The overall hybrid framework architecture has been illustrated in Fig.~\ref{fig0}. For ensembling, we have considered three different hybrid architectures where the prediction of each competing model is taken into consideration. The detailed descriptions of individual hybrid network architectures are in the following sub-sections.
 
 \subsection{Model 1(M1)}
 
 In this model \cite{Abani}, the images are resized to dimension $32 \times 32$ each having an RGB channel. The first part of the framework consists of classical CNN - a series of interlaced convolution layers to extract features and pooling layers to keep the dimensionality of the data at bay. At the first convolution layer, $10$ filters, each having dimension $5 \times 5$ have been used. The second convolution layer used involves $20$ filters of the same dimension as before. The RELU activation function is used for introducing non-linearity. After a couple of sets of convolution, Relu, and pooling, a dropout layer is introduced subsequently, in order to avoid overfitting. The CNN typically has two fully connected layers at the end, both having input sizes $500$. The output value of the last neuron in the fully connected CNN serves as input to the PQC in the form of the parameter value required by the parameterized circuit. The circuit measurement then serves as the final prediction for 0 or 1 as provided by a measurement in the z-basis $\sigma_z$.

 \subsection{Model 2(M2)}

 As in model 1, in the second model \cite{pyTorch} too, the RGB images are resized to dimension $32 \times 32$. The first convolution layer in the CNN involves $6$ filters having kernel size $5$. The second convolution layer used involves $16$ filters of the same dimension as before. The RELU activation function is used for introducing non-linearity. No dropout layer is included in this model. Three fully connected layers have been used. The first one transforms the input features into $120$ features. Thus, the fully connected last but one linear transformation layer has an input size of $120$ and an output size of $84$. The final fully connected layer transforms $84$ features into $1$, which in turn serves to be the input parameter to the quantum circuit. 

\subsection{Model 3(M3)}

The third network requires images to be resized to dimensions $250 \times 250$. The input image matrix is padded on all four sides, in this model and the stride of the convolution is considered $2$. Two layers of dropout are applied, one after each convolutional layer, to combat overfitting. Three fully connected layers combine to produce the final output of the classical part of the network. The layers respectively accept input features of sizes - $55815$, $120$, and $84$. The value of the last neuron is the feature value that is sent to the quantum circuit.

\subsection{Quantum Circuit}

\begin{figure}
\centering
\includegraphics[width=80mm]{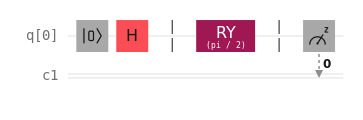}
\caption{Quantum Circuit used for Training the classifier} \label{fig1}
\end{figure}

The quantum circuit, in all three cases, comprises a Hadamard gate, followed by an $R_y$ rotation gate. The Hadamard gate performs the quantum-specific action of superposition of qubits and the $R_y$ gate is the PQC under consideration. The parameter $\theta$ of the PQC receives its value from the classical portion of the hybrid architecture. It is shown in Fig.~\ref{fig1}

 \section{Experiments}

 This section contains the description of the dataset used and details of the experimental procedure employed in this work.

 \subsection{Dataset Description}
\begin{figure}
\centering
 \begin{subfigure}[b]{\textwidth}
     \centering
     \includegraphics[width=\textwidth]{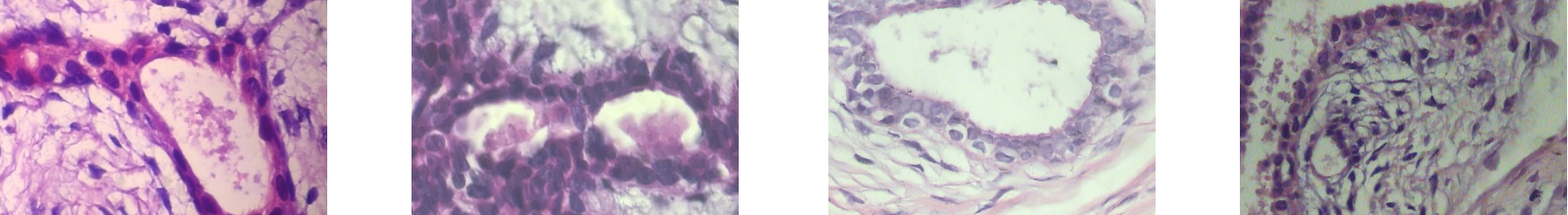}
     \caption{}
 \end{subfigure}
\hfill
 \begin{subfigure}[b]{\textwidth}
     \centering
     \includegraphics[width=\textwidth]{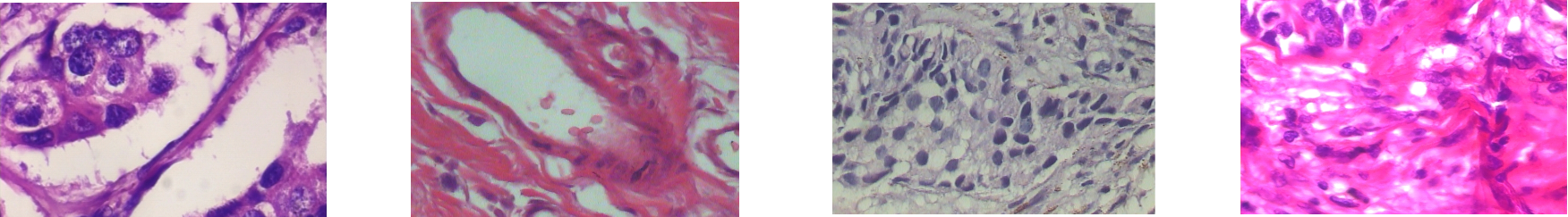}
     \caption{}
 \end{subfigure}
 \hfill
\caption{Images from BreakHis Dataset (a) Benign (b) Malignant} \label{fig4}
\end{figure}

 The study in this paper has been carried out on a breast cancer histopathology image dataset - BreakHis \cite{7312934}. Originally, BreakHis was a collection of 9,109 microscopic images of breast tumor tissue gathered from 82 patients using different magnifying scales (40X, 100X, 200X, and 400X). For our study, we have considered the problem of binary classification on the images recorded using a magnifying factor of 400X. The total set of images is broadly divided into two categories: benign tumors and malignant tumors. Out of the 2,480  benign and 5,429 malignant samples in the entire dataset, the aforementioned subset that we have selected contains 588 benign samples and 1232 malignant ones. Each image has a resolution of 700 $\times$ 460 pixels comprising 3-channel RGB with 8-bit depth in each channel. Fig.~\ref{fig4} shows a few sample images in the benign and malignant category.

 \subsection{Experimental Setup}

 The individual categories of benign and malignant are further subdivided into sets of data used for the purpose of training, validation, and testing the model in the ratio of 3:1:1.
A pytorch-qiskit pipeline was deployed to create the framework. For network training, we used the loss function Negative Log-Likelihood and the optimizer Adam. We use a quantum simulator to execute the quantum circuit operations. A total number of $100$ epochs are executed in order to perform the training phase, where the validation set is used to save the model displaying minimum loss.

  The entire classical-quantum framework in all three network models is trained as a whole,  end-to-end. This mitigates the requirement of any sort of pre-training. This derives the advantage of scalability and adaptability of the framework since the classical-quantum boundary can be adjusted depending on the availability of quantum resources.

  Training the network involved experimenting with the best permutations of the optimizer, learning rate, and loss function as hyperparameters in order to train over multiple epochs and validate the same. Hyperparameter tuning was carried out. The network was first trained on the training sample, it was then validated so as to determine whether the model suffered through under-fitting or over-fitting. For evaluating the performance of the networks, precision, recall and f-1 scores were used as metrics.

 \subsection{Ensemble}

The primary objective of designing multiple hybrid models is to achieve superior performance in the classification task at hand. Experimental results guide the selection of the most effective classifier. However, the frequent observation of non-overlapping misclassifications across different classifiers suggests that ensemble techniques could be employed to enhance overall classification performance. This is because different classifier designs may provide complementary insights into the patterns being classified. Ensemble learning leverages decisions from multiple models to arrive at a final, more accurate decision.

In this work, we focused on several well-established yet straightforward ensemble techniques based on aggregated predictions, including majority voting \cite{Kittler}, probability averaging \cite{Kuncheva}, and weighted probability averaging. The choice of these methods is primarily guided by their simple yet effective nature. In the weighted probability averaging method, the weight assigned to each classifier is inversely proportional to its number of misclassifications, thereby reducing the influence of models with poorer performance. To evaluate misclassifications, we considered instances where two other models correctly predicted a sample, while the model in question did not.

\section{Results}
In this section, we have briefly discussed the overall results and findings of our whole experiment.
\subsection{Performance of individual models}

\begin{figure}[h]
    \centering
    \includegraphics[width=\textwidth]{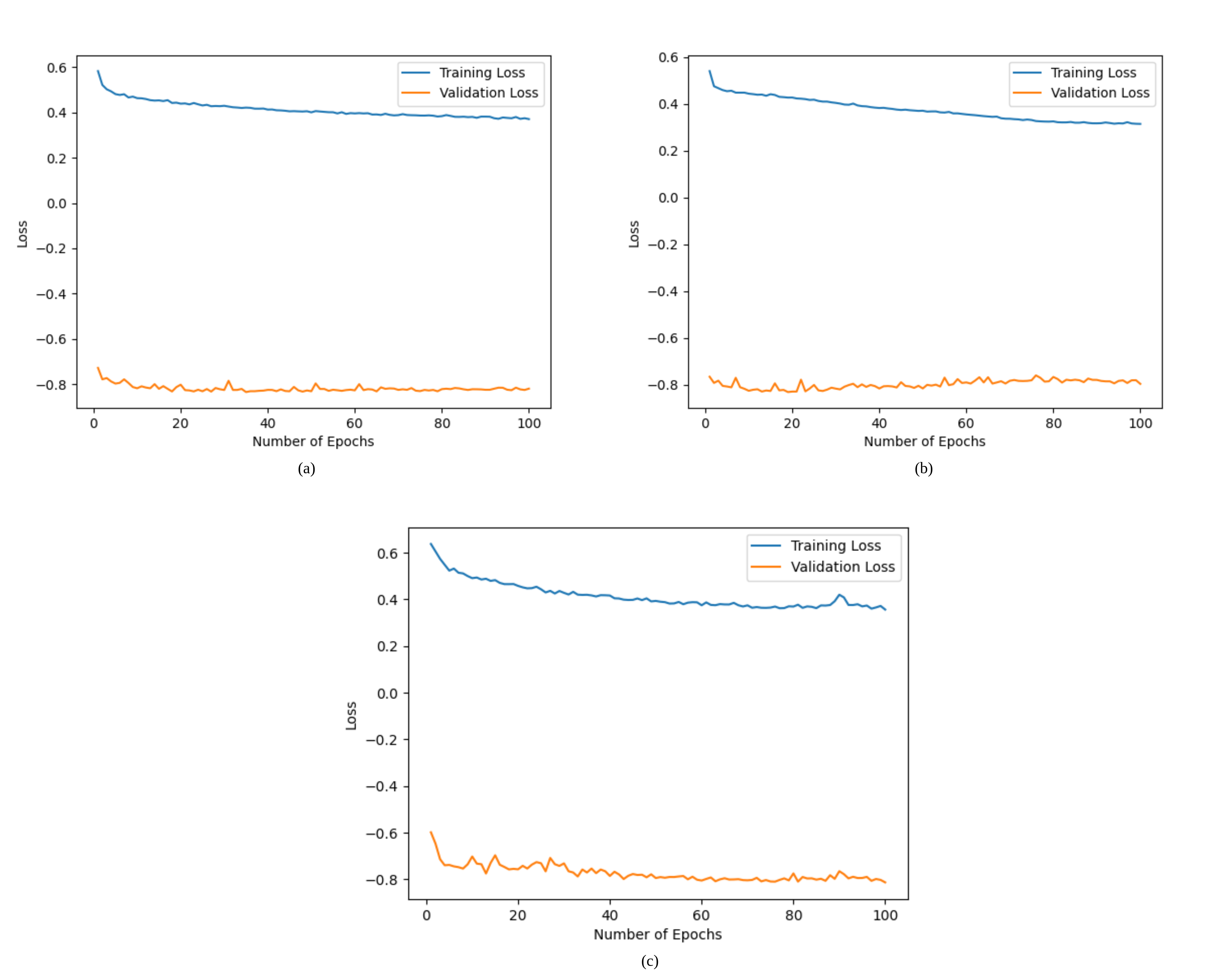}
    \caption{Training and Validation Loss curves of each individual model - (a) Model 1 (b) Model 2 (c) Model 3}
    \label{fig3}
\end{figure}

The training loss curves and validation loss curves have been depicted in Fig.~\ref{fig3}. The loss curves have maintained an optimal trend without leaving the suspicion of any possible underfitting and overfitting error.

\begin{figure}
    \centering
    \includegraphics[width=0.97\textwidth]{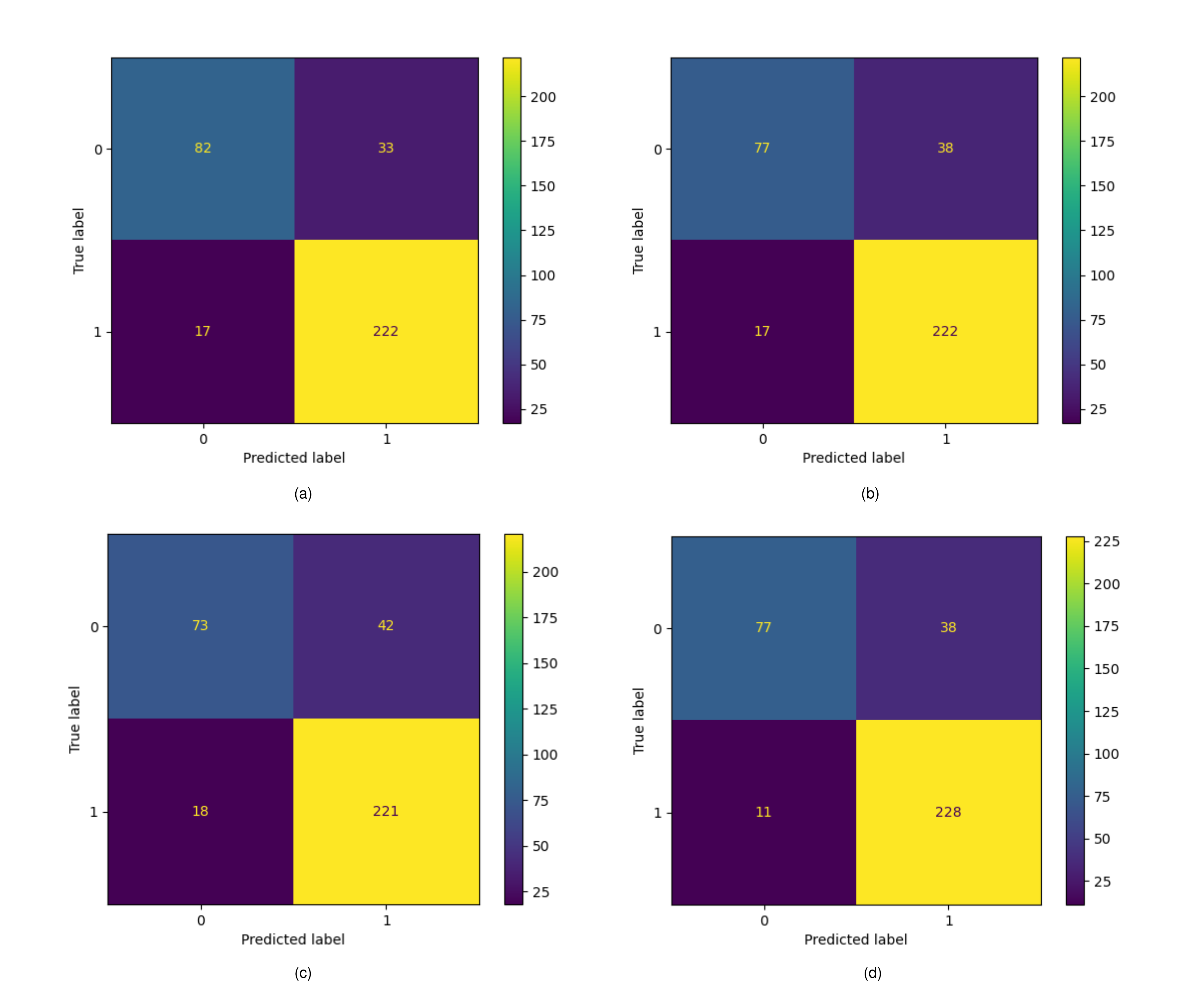}
    \caption{Confusion matrix of (a) Model 1 (b) Model 2 (c) Model 3 (d) Average Probability between model 2 and 3}
    \label{fig:2}
\end{figure}

Fig.~\ref{fig:2}(a), (b), (c), (d) shows the confusion matrices of each hybrid model. The testing accuracy of the first network amounts to $84.74\%$. Performance
is typically estimated on the basis of synthetic one-dimensional indicators such as precision, recall, or F1-score. The precision, recall, and F1-score of this model were respectively found as $84.85\%$, $79.45\%$, and $81.29\%$. Test accuracy of model 2 is obtained as $85.59\%$. The precision, recall, and F1-score of this model were respectively found as $85.85\%$, $80.53\%$, and $82.38\%$. For the third model, test accuracy is $85.59\%$. The respective precision, recall, and F1-score are $84.60\%$, $79.01\%$, and $80.89\%$.

 \subsection{Performance of Ensemble Techniques}
 Performance-wise, both the second and third models display similar accuracy values which incidentally is better than the first model. The ensemble modeling techniques aid in combining the models to generate superior classification performances.
 The performances of the individual base models and the ensemble techniques are shown in Table \ref{tab:my-table}. From Table.~\ref{tab:my-table} it is clear that the ensembling of all three base models improves the performance of the base models in terms of all four metrics. For example, majority voting among the three networks yields accuracy of $85.31\%$, precision of $85.9\%$, recall $79.87\%$, and F1-score of $81.88\%$. However, the Average Probability technique yields accuracy, precision, recall, and F1-score of $85.03\%$, $85.37\%$, $79.66\%$, and $81.58\%$ respectively. On the other hand, the Weighted Average technique achieves $85.88\%$ accuracy, $86.09\%$ precision, $81.17\%$ recall, and $82.78\%$ F1-score. 
 \begin{table}[h]
 \caption{Performances of all three individual Hybrid models and the ensemble techniques in Accuracy, Precision, Recall, and F1-score. The highest, second highest, and third highest are indicated by bold, underlined, and blue fonts respectively. }
\label{tab:my-table}
\resizebox{\textwidth}{!}{%
\begin{tabular}{c|c|cccc}
\hline
\textbf{} & \textbf{Model} & \textbf{Accuracy(\%)} & \textbf{Precision(\%)} & \textbf{Recall(\%)} & \textbf{F1-score(\%)} \\ \hline
\multirow{3}{*}{\begin{tabular}[c]{@{}c@{}}Individual\\  hybrid CNN\end{tabular}} & M1 & 84.74 & 84.85 & 79.45 & 81.29 \\ 
 & M2 & 85.59 & 85.85 & 80.53 & 82.38 \\ 
 & M3 & 85.59 & 84.60 & 79.01 & 80.89 \\ \hline
Majority Voting & M1 + M2 + M3 & 85.31 & 85.9 & 79.87 & 81.88 \\ \hline
\multirow{4}{*}{\begin{tabular}[c]{@{}c@{}}Average \\ Probability\end{tabular}} & M1 + M2 & \underline{86.16} & \textbf{86.61} & \underline{81.17} & \underline{83.08} \\ 
 & M2 + M3 & \textbf{86.72} & \textcolor{blue}{86.54} & \textbf{82.49} & \textbf{84.04} \\ 
 & M1 + M3 & 85.03 & 85.09 & 79.88 & 81.69 \\ 
 & M1 + M2 + M3 & 85.03 & 85.37 & 79.66 & 81.58 \\ \hline
\multirow{4}{*}{\begin{tabular}[c]{@{}c@{}}Weighted \\ Average\end{tabular}} & M1 + M2 & \textcolor{blue}{86.15} & \underline{86.60} & \underline{81.17} & \textcolor{blue}{83.07} \\ 
 & M2 + M3 & 85.59 & 85.58 & 80.75 & 82.49 \\ 
 & M1 + M3 & 84.75 & 85.09 & 79.88 & 81.69 \\ 
 & M1 + M2 + M3 & 85.88 & 86.09 & \textcolor{blue}{81.17} & 82.78 \\ \hline
\end{tabular}%
}
\end{table}
 We have experimented taking two models at a time as well. The Average Probability technique between Model 2 and Model 3 achieves $86.72\%$ accuracy, $82.49\%$ recall, and $84.04\%$ F1-score, which is the best performance in the respective category. A reason for that is, both models 2 and 3 have similar performances which is evident again from Table~\ref{tab:my-table}. The Average Probability also performs good on combining models 1 and 2. For this combination, the precision value is the highest, viz, $86.61\%$. We have shown the confusion matrix of the best performing model, i.e., combination of models 2 and 3 with Average Probability, in Fig.~\ref{fig:2}(d). The weighted average procedure too indicates fair performance figures on the combination of the first two models. Thus, from the values tabulated, it is evident that, fusing the individual hybrid models using ensembling, results in $1.13\%$ increase in accuracy, and almost $2\%$ increase in recall and F1-score.

\section{Conclusion}

In this paper, we have considered three hybrid network architectures based on classical convolutional neural networks enhanced using parameterized quantum circuits that perform the task of binary classification of histopathological images. The first model serves as the base architecture which we fine-tune to produce the two other models. The three models are further ensembled using different ensembling techniques wherein the average probability as well as the weighted average of a combination of models yield better performance than the individual ones.

The experiments we carried out involved quantum simulators executed on classical hardware. In such hardware, quantum neural networks do not offer any advantage in training time over classical convolutional neural networks. In fact, classical CNNs actually exhibit shorter training times than QNNs. This increased training time for QNNs can be attributed to quantum state preparation, preparing the quantum circuit involving multiple gates, and execution of the gates. The time complexity is proportional to the circuit complexity, and hence involving further layers of variational quantum components would entail greater training time. Hence, the simplistic model is considered in this work. However, current state-of-the-art network architectures need to be explored in place of the simplistic architectures described here. Also, the task of multi-level classification remains to be explored in the future.


\subsubsection{Acknowledgements} The authors are thankful to CMATER Lab, CSE Department, Jadavpur University for providing the infrastructural support during the experiments. We also acknowledge the invaluable insights provided by Mr. Soumyajyoti Dey, research scholar, Jadavpur University.

\begin{credits}


\end{credits}
%
%
%
%

\frenchspacing
\bibliographystyle{plain}
\bibliography{References.bib}

\end{document}